\documentclass[11pt,a4paper]{article}
\usepackage{emnlp2018} %
\usepackage{times}
\usepackage{latexsym}
\usepackage[normalem]{ulem}

\aclfinalcopy %

\title{Instructions for ACL 2018 Proceedings}

\author{First Author \\
  Affiliation / Address line 1 \\
  Affiliation / Address line 2 \\
  Affiliation / Address line 3 \\
  {\tt email@domain} \\\And
  Second Author \\
  Affiliation / Address line 1 \\
  Affiliation / Address line 2 \\
  Affiliation / Address line 3 \\
  {\tt email@domain} \\}

\usepackage{scrextend}

\usepackage{amssymb}
\usepackage{amsmath}
\usepackage{dsfont}
\usepackage{graphicx}
\usepackage{multirow}
\usepackage{tikz,pgfplots}
\usepgfplotslibrary{groupplots}
\usepackage{balance}
\usepackage{xspace}
\usepackage{breqn}
\usepackage{mdframed}

\usepackage{amsthm}

\newtheorem*{definition*}{Definition}
\newtheorem*{theorem*}{Theorem}

\usepackage{titlesec}
\titlespacing*{\section} {0pt}{1.2ex plus 1ex minus .2ex}{0.8ex plus .2ex}
\titlespacing*{\paragraph} {0pt}{0.2ex plus 0.05ex minus .05ex}{1em}

\usepackage{caption}
\DeclareCaptionFormat{capfont}{\fontsize{10}{6}\selectfont#1#2#3}
\usepackage{subcaption}
\makeatletter
\g@addto@macro\small{%
  \setlength\abovedisplayskip{-5pt}
  \setlength\abovedisplayshortskip{-5pt}
  \setlength\belowdisplayshortskip{-7pt}
  \setlength\belowdisplayskip{-7pt}
}
\makeatother

\usepackage{array}
\newcolumntype{L}[1]{>{\raggedright\let\newline\\\arraybackslash\hspace{0pt}}m{#1}}
\newcolumntype{C}[1]{>{\centering\let\newline\\\arraybackslash\hspace{0pt}}m{#1}}
\newcolumntype{R}[1]{>{\raggedleft\let\newline\\\arraybackslash\hspace{0pt}}m{#1}}

\date{}

\usepackage{algorithmicx} 
\usepackage[noend]{algpseudocode}
\usepackage{algorithm}
\usepackage[T1]{fontenc}

\algnewcommand\algorithmicdefinitions{\textbf{Definitions:}}
\algnewcommand\Definitions{\item[\algorithmicdefinitions]}
\renewcommand{\algorithmiccomment}[1]{{\color{gray}\raisebox{1px}{\texttt{\guillemotright}} #1}}
\algnewcommand{\LineComment}[1]{\Statex \hskip\ALG@thistlm \algorithmiccomment{#1}}
\algrenewcommand\alglinenumber[1]{\footnotesize #1:}
\algrenewcommand\algorithmicindent{1.0em}%
\makeatletter
\newcommand{\StatexIndent}[1][3]{%
  \setlength\@tempdima{\algorithmicindent}%
  \Statex\hskip\dimexpr#1\@tempdima\relax}

\usepackage{arydshln}
\newcommand{\dline}{\hdashline[0.5pt/1pt]}

\newcommand{\eat}[1]{}

\newcommand{\nlstring}[1]{{\em #1}}

\newcommand{\system}[1]{\textsc{#1}}

\newcommand{\state}{s}
\newcommand{\allstate}{\mathcal{S}}
\newcommand{\startstate}{s_{1}}
\newcommand{\goalstate}{s_g}

\newcommand{\instruction}{\bar{x}}
\newcommand{\allinstruction}{\mathcal{X}}
\newcommand{\token}{x}
\newcommand{\action}{a}
\newcommand{\allaction}{\mathcal{A}}

\newcommand{\image}{\mathbf{I}}

\newcommand{\act}[1]{{\tt \MakeUppercase{#1}}}
\newcommand{\stopaction}{\act{Stop}}

\newcommand{\pose}{p}
\newcommand{\goalpos}{l_g}

\newcommand{\pano}{\image_P}

\newcommand{\lingunet}{\textsc{LingUNet}\xspace}
\newcommand{\unet}{\textsc{U-Net}\xspace}
\newcommand{\conv}{\textsc{CNN}}
\newcommand{\lstm}{\textsc{LSTM}}
\newcommand{\lstmrep}{\mathbf{l}}
\newcommand{\langrep}{\mathbf{\instruction}}

\newcommand{\ostate}{\tilde{s}}

\newcommand{\param}{\theta}
\newcommand{\embed}{\psi}

\newcommand{\attnmask}{\mathbf{M}}
\newcommand{\outofsightflag}{o}

\newcommand{\featmap}{\mathbf{F}}
\newcommand{\featmaptxt}{\mathbf{G}}
\newcommand{\featmapdeconv}{\mathbf{H}}

\newcommand{\posemb}{\featmap^p}
\newcommand{\goalprob}{P_g}
\newcommand{\goalprobbias}{b_g}
\newcommand{\kernel}{\mathbf{K}}
\newcommand{\affine}{\textsc{Affine}}
\newcommand{\convolve}{\textsc{Convolve}}
\newcommand{\dropout}{\textsc{Dropout}}
\newcommand{\deconv}{\textsc{Deconv}}

\newcommand{\softmax}{\textsc{Softmax}}

\newcommand{\acthidden}{y}

\newcommand{\reward}{R}

\newcommand{\navdrone}{\textsc{Lani}\xspace}
\newcommand{\house}{\textsc{Chai}\xspace}
\newcommand{\chalet}{\textsc{Chalet}\xspace}

\definecolor{drone-green}{RGB}{10,255,0}
\definecolor{drone-pink}{RGB}{250,7,255}
\definecolor{drone-yellow}{RGB}{255,200,20}
\definecolor{drone-cyan}{RGB}{5,200,255}

\newcommand{\coloruline}[1]{\bgroup\markoverwith{\textcolor{#1}{\rule[-0.5ex]{2pt}{1pt}}}\ULon}

\newcommand\dronegreenuline{\coloruline{drone-green}}
\newcommand\dronepinkuline{\coloruline{drone-pink}}
\newcommand\droneyellowuline{\coloruline{drone-yellow}}
\newcommand\dronecyanuline{\coloruline{drone-cyan}}

\newcommand{\startarrow}[2]{
\begin{tikzpicture}[overlay]
  \begin{scope}[shift={(#1,#2)}]
    \tikzstyle{arrowfill} = [fill=white, draw=black]
    \filldraw[arrowfill](0.2,0)--(0.2,0.2)--(0.4,0.2)--(0,0.35)--(-0.4,0.2)--(-0.2,0.2)--(-0.2,0)--cycle;
  \end{scope}
\end{tikzpicture}
}

\newcommand{\textoverlay}[3]{
\begin{tikzpicture}[overlay]
  \begin{scope}[shift={(#1,#2)}]
    \node[draw=white] at (0,0) {\color{white}#3};
  \end{scope}
\end{tikzpicture}
}

\title{Mapping Instructions to Actions in 3D Environments\\ with Visual Goal Prediction}%

\author{
\textbf{
Dipendra Misra  \hspace{5mm} Andrew Bennett \hspace{5mm} Valts Blukis} \\
\textbf{
Eyvind Niklasson \hspace{5mm} Max Shatkhin \hspace{5mm} Yoav Artzi} \vspace{2mm} \\ 
Department of Computer Science and Cornell Tech, 
Cornell University,
New York, NY, 10044  \\
\texttt{\{dkm, awbennett, valts, yoav\}@cs.cornell.edu} \\ \texttt{\{een7, ms3448\}@cornell.edu}
}

\begin{document}
\maketitle
\begin{abstract}
We propose to decompose instruction execution to goal prediction and action generation. 
We design a model that maps raw visual observations to goals using \lingunet, a language-conditioned image generation network, and then generates the actions required to complete them. 
Our model is trained from demonstration only without external resources. 
To evaluate our approach, we introduce two  benchmarks for instruction following: \navdrone, a navigation task; and \house, where an agent executes household instructions. 
Our evaluation demonstrates the advantages of our model decomposition, and illustrates the challenges posed by our new benchmarks.

\end{abstract}

\section{Introduction}
\label{sec:intro}

Executing instructions in interactive environments requires mapping natural language and  observations to actions. 
Recent approaches propose learning to directly map from inputs to actions, for example given language and either structured observations~\cite{Mei:16neuralnavi,Suhr:18situated} or raw visual observations~\cite{Misra:17instructions,Xiong:18scheduled}. 
Rather than using a combination of models, these approaches learn a single model to solve language, perception, and planning challenges. 
This reduces the amount of engineering required and eliminates the need for hand-crafted meaning representations. 
At each step, the agent maps its current inputs to the next action using a single learned function that is executed repeatedly until task completion.

\begin{figure}[t]
\centering
  \begin{minipage}{0.49\linewidth}
    \centering
    \frame{\includegraphics[width=0.92\textwidth]{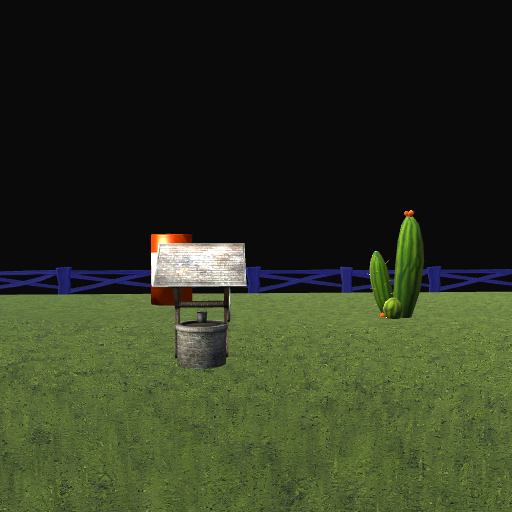}}  \\[3pt]
      \footnotesize
      \begin{tabular}{|p{0.885\textwidth}|}
      \hline
      \nlstring{After reaching the hydrant head towards the blue fence and pass towards the right side of the well.} \\
      \hline
      \end{tabular}
  \end{minipage}~
  \begin{minipage}{0.49\linewidth}
    \centering
    \frame{\includegraphics[width=0.92\textwidth]{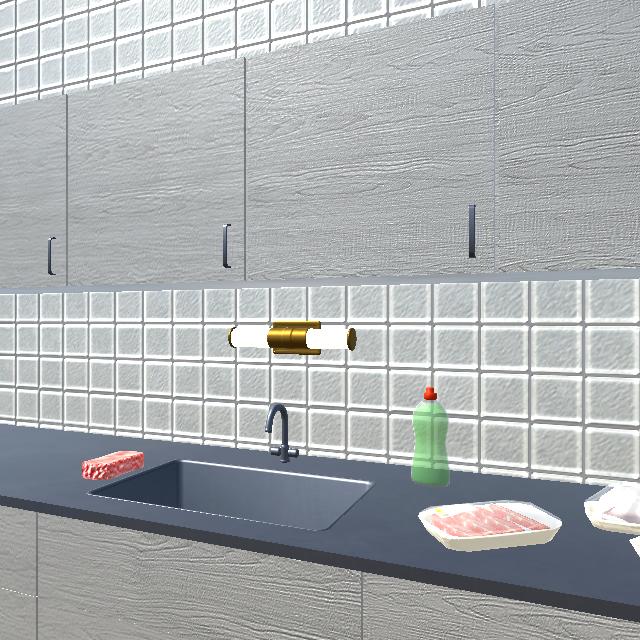}}  \\[3pt]
      \footnotesize
      \begin{tabular}{|p{0.885\textwidth}|}
      \hline
      \nlstring{Put the cereal, the sponge, and the dishwashing soap into the cupboard above the sink.} \\
      \hline
      \end{tabular}
  \end{minipage}
  \caption{Example instructions from our two tasks: \navdrone (left) and \house (right). \navdrone is a landmark navigation task, and \house is a corpus of instructions in the \chalet environment.}
  \label{fig:examples}
\end{figure}

Although executing the same computation at each step simplifies modeling, it exemplifies  certain inefficiencies; while the agent needs to decide what action to take at each step, identifying its goal  is only required once every several steps or even once per execution. 
The left instruction in Figure~\ref{fig:examples} illustrates this. 
The agent can compute its goal once given the initial observation, and given this goal can then generate the actions required. 
In this paper, we study a new model that explicitly distinguishes between goal selection and action generation, and introduce two instruction following benchmark tasks to evaluate it.

Our model decomposes into goal prediction and action generation. 
Given a natural language instruction and system observations, the model predicts the  goal to complete. 
Given the goal, the model generates a sequence of actions.

The key challenge we address is designing the goal representation. 
We avoid manually designing a meaning representation, and predict the goal in the agent's observation space. 
Given the image of the environment the agent observes, we generate a probability distribution over the image to highlight the goal location. 
We treat this prediction as image generation, and develop \lingunet, a language conditioned variant of the \unet image-to-image architecture~\cite{ronneberger2015u}. 
Given the visual goal prediction, we generate actions using a recurrent neural network (RNN).

Our model decomposition offers two key advantages. 
First, we can use different learning methods as appropriate for the goal prediction and action generation problems. 
We find supervised learning more effective for goal prediction, where only a limited amount of natural language data is available. 
For action generation, where exploration is critical, we use policy gradient in a contextual bandit setting~\cite{Misra:17instructions}. 
Second, the goal distribution is easily interpretable by overlaying it on the agent observations. 
This can be used to increase the safety of physical systems by letting the user verify the goal before any action is executed.
Despite the decomposition, our approach retains the advantages of the single-model approach. 
It does not require designing intermediate representations, and training does not rely on external resources, such as pre-trained parsers or object detectors, instead using  demonstrations only. 

We introduce two new benchmark tasks with different levels of complexity of goal prediction and action generation. 
\navdrone is a 3D navigation environment and corpus, where an agent navigates between landmarks. The corpus includes 6,000 sequences of natural language instructions, each containing on average 4.7 instructions.  
\house is a corpus of 1,596 instruction sequences, each including 7.7 instructions on average, for \chalet, a 3D house environment~\cite{yan2018chalet}. Instructions combine navigation and simple manipulation, including moving objects and opening containers. 
Both tasks require solving language challenges, including spatial and temporal reasoning, as well as complex perception and planning problems. 
While \navdrone provides a task where most instructions include a single goal, the \house instructions often require multiple intermediate goals. 
For example, the household instruction in Figure~\ref{fig:examples} can be decomposed to eight goals: opening the cupboard, picking each item and moving it to the cupboard, and  closing the cupboard. 
Achieving each goal requires multiple actions of different types, including moving and acting on objects. 
This allows us to experiment with a simple variation of our model to generate intermediate goals.

We compare our approach to multiple recent methods. 
Experiments on the \navdrone navigation task indicate that decomposing goal prediction and action generation significantly improves instruction execution performance. 
While we observe similar trends on the \house instructions, results are overall weaker, illustrating the complexity of the task. 
We also observe that inherent ambiguities in instruction following make exact goal identification difficult, as demonstrated by imperfect human performance. 
However, the gap to human-level performance still remains large across both tasks. 
Our code and data are available at \href{https://github.com/clic-lab/ciff}{{\tt github.com/clic-lab/ciff}}.

\section{Technical Overview}
\label{sec:overview}
\paragraph{Task} Let $\allinstruction$ be the set of all \emph{instructions}, $\allstate$ the set of all \emph{world states}, and $\allaction$ the set of all \emph{actions}. 
An instruction $\instruction \in \allinstruction$ is a sequence $\langle \token_1, \dots, \token_n \rangle$, where each $\token_i$ is a token. 
The agent executes instructions by generating a sequence of actions, and indicates execution completion with the special action $\stopaction$.

The sets of actions $\allaction$ and states $\allstate$ are domain specific. 
In the navigation domain \navdrone, the actions include moving the agent and changing its orientation. The state information includes the position and orientation of the agent and the different landmarks. The agent actions in the \chalet house environment include moving and changing the agent orientation, as well as an object interaction action. 
The state encodes the position and orientation of the agent and all objects in the house. For interactive objects, the state also includes their status, for example if a drawer is open or closed. 
In both domains, the actions are discrete. 
The domains are described in Section~\ref{sec:data}.

\paragraph{Model} 
The agent does not observe the world state directly, but instead observes its pose and an RGB image of the environment from its point of view. 
We define these observations as the agent context $\ostate$. 
An agent model is a function from an agent context $\ostate$ to an action $\action \in \allaction$. 
We model goal prediction as predicting a probability distribution over the agent visual observations, representing the likelihood of locations or objects in the environment being target positions or objects to be acted on. 
Our model is described in Section~\ref{sec:model}.

\paragraph{Learning} We assume access to training data with $N$ examples $\{ (\instruction^{(i)}, \startstate^{(i)},  \goalstate^{(i)})\}_{i=1}^N$, where $\instruction^{(i)}$ is an instruction, $\startstate^{(i)}$ is a start state, and $\goalstate^{(i)}$ is the goal state. 
We decompose learning; training goal prediction using supervised learning, and action generation using oracle goals with policy gradient in a contextual bandit setting.
We assume an instrumented environment with access to the world state, which is used to compute rewards during training only. 
Learning is described in Section~\ref{sec:learn}.

\paragraph{Evaluation} We evaluate task performance on a test set $\{(\instruction^{(i)}, \startstate^{(i)}, \goalstate^{(i)}) \}^M_{i=1}$, where $\instruction^{(i)}$ is an instruction, $\startstate^{(i)}$ is a start state, and $\goalstate^{(i)}$ is the goal state. We evaluate task completion accuracy and the distance of the agent's final state to $\goalstate^{(i)}$. 

\section{Related Work}
\label{sec:related}

Mapping instruction to action has been studied extensively with intermediate symbolic representations~\cite[e.g.,][]{Chen:11,Kim:12,Artzi:13,Artzi:14,Misra:15highlevel,Misra:16telldave}. 
Recently, there has been growing interest in direct mapping from raw visual observations to actions~\cite{Misra:17instructions,Xiong:18scheduled,mattersim,Fried:18r2r}. 
We propose a model that enjoys the benefits of such direct mapping, but explicitly decomposes that task to interpretable goal prediction and action generation. 
While we focus on natural language, the problem has also been studied using synthetic language~\cite{chaplot2017gated,Hermann2017}. 

Our model design is related to hierarchical reinforcement learning, where sub-policies at different levels of the hierarchy are used at different frequencies~\cite{sutton1998intra}. 
\citet{Oh:17ZeroShotTG} uses a two-level hierarchy for mapping synthetic language to actions. 
Unlike our visual goal representation, they use an opaque vector representation. 
Also, instead of reinforcement learning, our methods emphasize sample efficiency.

Goal prediction is related to referring expression interpretation~\cite{Matuszek:12,Krishnamurthy:13,Kazemzadeh2014ReferItGameRT,Kong2014WhatAY,Yu2016ModelingCI,Mao2016,kitaev2017misty}. 
While our model solves a similar problem for goal prediction, we focus on detecting visual goals for actions, including both navigation and manipulation, as part of an instruction following model. 
Using formal goal representation for instruction following was studied by \citet{MacGlashan2015GroundingEC}. 
In contrast, our model generates a probability distribution over images, and  does not require an ontology.

Our data collection is related to existing work. 
\navdrone is inspired by the HCRC Map Task~\cite{Anderson:91}, where  a leader directs a follower to navigate between landmarks on a map. 
We use a similar task, but our scalable data collection process allows for a significantly larger corpus. We also provide an interactive navigation environment, instead of only map diagrams. 
Unlike Map Task, our leaders and followers do not interact in real time. This abstracts away interaction challenges, similar to how the SAIL navigation corpus was collected~\cite{MacMahon:06}. 
\house instructions were collected using scenarios given to workers, similar to the ATIS collection process~\cite{Hemphill:90atis,Dahl:94}. 
Recently, multiple 3D research environments were released. 
\navdrone has a significantly larger state space than existing navigation environments~\cite{Hermann2017,chaplot2017gated}, and \chalet, the environment used for \house, is larger and has more complex manipulation compared to similar environments~\cite{gordon2018iqa,das2018embodied}. 
In addition, only synthetic language data has been released for these environment.  
An exception is the Room-to-Room dataset~\cite{mattersim} that makes use of an environment of connected panoramas of house settings. Although it provides a realistic vision challenge, unlike our environments, the state space is limited to a small number of panoramas and manipulation is not possible.

\section{Model}
\label{sec:model}

\begin{figure*}
  \begin{center}
  \includegraphics[width=0.8\textwidth,clip,trim=2 133 3 126]{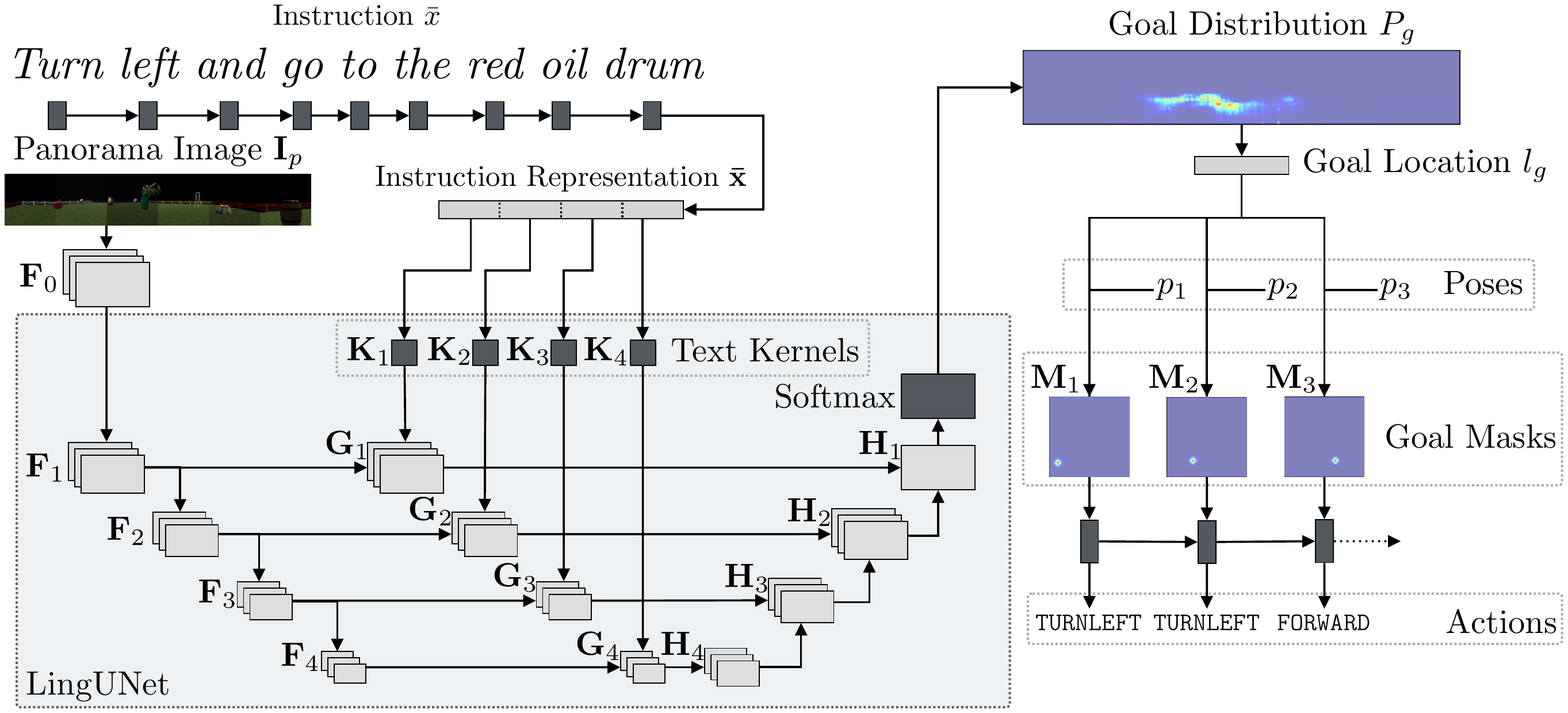}
  \end{center}
  \caption{An illustration for our architecture (Section~\ref{sec:model}) for the instruction \nlstring{turn left and go to the red oil drum} with a \lingunet depth of $m=4$. The instruction $\instruction$ is mapped to $\langrep$ with an RNN, and the initial panorama  observation  $\pano$ to $\featmap_0$ with a CNN. \lingunet generates $\featmapdeconv_1$, a visual representation of the goal. First, a sequence of convolutions maps the image features $\featmap_0$ to feature maps $\featmap_1$,\dots,$\featmap_4$. The text representation $\langrep$ is used to generate the kernels $\kernel_1$,\dots,$\kernel_4$, which are convolved to generate the text-conditioned feature maps $\featmaptxt_1$,\dots,$\featmaptxt_4$. These feature maps are de-convolved to $\featmapdeconv_1$,\dots,$\featmapdeconv_4$. The goal probability distribution $\goalprob$ is computed from $\featmapdeconv_1$. The goal location is the inferred from the max of $\goalprob$. Given $\goalpos$ and $\pose_t$, the  pose at step $t$, the goal mask $\attnmask_t$ is computed and passed into an RNN that outputs the action to execute.}
  \label{fig:model}
\end{figure*}

We model the agent policy as a neural network. The agent observes the world state $\state_t$ at time $t$ as an RGB image $\image_t$. 
The agent context $\ostate_t$, the information available to the agent to select the next action $\action_t$, is a tuple $(\instruction, \pano, \langle (\image_1, \pose_1),\dots,(\image_t,\pose_t) \rangle )$, where $\instruction$ is the natural language instructions, $\pano$ is a panoramic view of the environment from the starting position at time $t=1$, and $\langle (\image_1, \pose_1),\dots,(\image_t,\pose_t) \rangle$ is the sequence of observations $\image_t$ and poses $\pose_t$ up to time $t$. 
The panorama $\pano$ is generated through deterministic exploration by rotating $360^\circ$ to observe the environment at the beginning of the execution.\footnote{The panorama is a concatenation of deterministic observations along the width dimension. For simplicity, we do not include these deterministic steps in the execution.} 

The model includes two main components: goal prediction and action generation. 
The agent uses the panorama $\pano$ to predict the goal location $\goalpos$. At each time step $t$, a projection of the goal location into the agent's current view $\attnmask_t$  is given as input to an RNN to generate actions. 
The probability of an action $\action_t$ at time $t$ decomposes to:

\begin{small}
\begin{align*}
  P(\action_t \mid \ostate_t) = \sum_{l_g} \Big( &P(\goalpos \mid \instruction, \pano) \nonumber \\[-5pt]
                                                   &P(\action_t \mid \goalpos, (\image_1,\pose_1), \dots,(\image_{t}, \pose_{t}) ) \Big)\;\;,
\end{align*}
\end{small}

\noindent
where the first term puts the complete distribution mass on a single location (i.e., a delta function). 
Figure~\ref{fig:model} illustrates the model.

\paragraph{Goal Prediction}

To predict the goal location, we generate a probability distribution $\goalprob$ over a feature map $\featmap_0$ generated using convolutions from the initial panorama observation $\pano$. 
Each element in the probability distribution $\goalprob$ corresponds to an area in $\pano$. 
Given the instruction $\instruction$ and panorama $\pano$, we first generate their representations. 
From the panorama $\pano$, we generate a feature map $\featmap_0 = [\conv_0(\pano);\posemb]$, 
where $\conv_0$ is a two-layer convolutional neural network~\cite[CNN;][]{lecun1998gradient} with rectified linear units~\cite[ReLU;][]{nair2010rectified} and $\posemb$ are positional embeddings.\footnote{We generate $\posemb$ by creating a channel for each  deterministic observation used to create the panorama, and setting all the pixels corresponding to that observation location in the panorama to $1$ and all others to  $0$. The number of observations depends on the agent's camera angle.} 
The concatenation is along the channel dimension. 
The instruction $\instruction = \langle \token_1, \cdots \token_n\rangle$ is mapped to a sequence of hidden states $\lstmrep_i = \lstm_x(\embed_x(\token_i), \lstmrep_{i-1})$, $i=1,\dots,n$ using a learned embedding function   $\embed_x$ and  a long short-term memory~\cite[LSTM;][]{Hochreiter:97lstm} RNN $\lstm_x$. 
The instruction representation is $\langrep = \lstmrep_n$.

We generate the probability distribution $\goalprob$ over pixels in $\featmap_0$ using \lingunet. 
The architecture of \lingunet is inspired by the \unet image generation method~\cite{ronneberger2015u}, except that the reconstruction phase is conditioned on the natural language instruction.
\lingunet first applies $m$ convolutional layers to generate a sequence of feature maps $\featmap_j= \conv_j(\featmap_{j-1})$, $j = 1\dots m$, 
where each $\conv_j$ is a convolutional layer with leaky ReLU non-linearities~\cite{maas2013rectifier} and instance normalization~\cite{ulyanov1607instance}. 
The instruction representation $\langrep$ is split evenly into $m$ vectors $\{\langrep_j\}_{j=1}^m$, each is used to create a $1\times 1$ kernel $\kernel_{j} = \affine_j(\langrep_j)$, where each $\affine_j$ is an affine transformation followed by normalizing and reshaping.  
For each $\featmap_j$, we apply a 2D $1\times 1$ convolution using the text kernel $\kernel_j$ to generate a text-conditioned feature map $\featmaptxt_j = \convolve(\kernel_j, \featmap_j)$, where $\convolve$ convolves the kernel over the feature map. 
We then perform $m$ deconvolutions to generate a sequence of feature maps $\featmapdeconv_m$,\dots,$\featmapdeconv_1$: 

\begin{small}
\begin{eqnarray*}
\featmapdeconv_m &=& \deconv_m(\dropout(\featmaptxt_m))\\
\featmapdeconv_j &=& \deconv_j([\featmapdeconv_{j+1}; \featmaptxt_j])\,\,.
\end{eqnarray*}
\end{small}

\noindent
$\dropout$ is dropout regularization~\cite{srivastava2014dropout} and each $\deconv_j$ is a deconvolution operation followed a leaky ReLU non-linearity and instance norm.\footnote{$\deconv_1$ does deconvolution only.}
Finally, we generate $\goalprob$ by applying a softmax to $\featmapdeconv_1$ and an additional learned scalar bias term $\goalprobbias$ to represent events where the goal is out of sight. For example, when the agent already stands in the goal position and therefore the panorama does not show it. 

We use $\goalprob$ to predict the goal position in the environment. 
We first select the goal pixel in $\featmap_0$ as the pixel corresponding to the highest probability element in $\goalprob$. 
We then identify the corresponding 3D location $\goalpos$ in the environment  using backward camera projection, which is computed given the camera parameters and  $\pose_1$, the agent pose at the beginning of the execution.

\paragraph{Action Generation}
Given the predicted goal $\goalpos$, we generate actions using an RNN. 
At each time step $t$, given $\pose_t$, we generate the goal mask $\attnmask_t$, which has the same shape as the observed image $\image_t$. The goal mask $\attnmask_t$ has a value of $1$ for each element that corresponds to the goal location $\goalpos$ in $\image_t$. We do not distinguish between visible or occluded locations. All other elements are set to $0$. We also maintain an out-of-sight flag $\outofsightflag_t$ that is set to $1$ if (a) $\goalpos$ is not within the agent's view; or (b) the max scoring element in $\goalprob$ corresponds to $\goalprobbias$, the term for events when the goal is not visible in $\pano$. Otherwise, $\outofsightflag_t$ is set to $0$. We compute an action generation hidden state $\acthidden_t$ with an RNN: 

\begin{small}
\begin{equation}
\nonumber \acthidden_t = \lstm_A \left( \affine_A([\textsc{Flat}(\attnmask_t);\outofsightflag_t ]), \acthidden_{t-1} \right)\;\;,
\end{equation}  
\end{small}

\noindent
where $\textsc{Flat}$ flattens  $\attnmask_t$ into a vector, $\affine_A$ is a learned affine transformation with ReLU, and $\lstm_A$ is an LSTM RNN. 
The previous hidden state $\acthidden_{t-1}$ was computed when generating the previous action, and the RNN is extended gradually during execution. 
Finally, we compute a probability distribution over actions:

\begin{small}
\begin{eqnarray*}
\nonumber  P(\action_t \mid \goalpos, (\image_1,\pose_1), \dots,(\image_{t}, \pose_{t}) ) &=& \\ && \hspace{-10em} \softmax(\affine_p([\acthidden_t; \embed_T(t)]) )\;\;,
\end{eqnarray*}
\end{small}

\noindent
where $\embed_T$ is a learned embedding lookup table for the current time~\cite{chaplot2017gated} and $\affine_p$ is a learned affine transformation. 

\paragraph{Model Parameters}

The model parameters $\param$ include the parameters of the convolutions $\conv_0$ and the components of \lingunet: $\conv_j$, $\affine_j$, and $\deconv_j$ for $j=1,\dots,m$. In addition we learn two affine transformations $\affine_A$ and $\affine_p$, two RNNs $\lstm_x$ and $\lstm_A$, two embedding functions $\embed_x$ and $\embed_T$, and the goal distribution bias term $\goalprobbias$. 
In our experiments (Section~\ref{sec:experiments}), all parameters are learned without external resources.

\section{Learning}
\label{sec:learn}

Our modeling decomposition enables us to choose different learning algorithms for the two parts. 
While reinforcement learning is commonly deployed for tasks that benefit from exploration~\cite[e.g.,][]{peters2008reinforcement,Mnih:13atari}, these methods require many samples due to their high sample complexity. 
However, when learning with natural language, only a  relatively small number of samples is realistically available. 
This problem was addressed in prior work by learning in a contextual bandit setting~\cite{Misra:17instructions} or mixing reinforcement and supervised learning~\cite{Xiong:18scheduled}. 
Our decomposition uniquely offers to tease apart the language understanding problem and address it with supervised learning, which generally has lower sample complexity. 
For action generation though, where exploration can be autonomous, we use policy gradient in a contextual bandit setting~\cite{Misra:17instructions}.

We assume access to training data with $N$ examples $\{ (\instruction^{(i)}, \startstate^{(i)},  \goalstate^{(i)})\}_{i=1}^N$, where $\instruction^{(i)}$ is an instruction, $\startstate^{(i)}$ is a start state, and $\goalstate^{(i)}$ is the goal state. 
We train the goal prediction component by minimizing the cross-entropy of the predicted distribution with the gold-standard goal distribution. The gold-standard goal distribution is a deterministic distribution with probability one at the pixel corresponding to the goal location if the goal is in the field of view, or probability one at the extra out-of-sight position otherwise. The gold location is the agent's location in $\goalstate^{(i)}$. We update the model parameters using Adam~\cite{Kingma:14adam}.

We train action generation by maximizing the expected immediate reward the agent observes while exploring the environment.  
The objective for a single example $i$ and time stamp $t$ is: 

\begin{small}
\begin{equation}
\nonumber J = \sum_{a\in \allaction} \pi(\action\mid \ostate_t) R^{(i)}(\state_t, \action) + \lambda H(\pi(. \mid \ostate_t))\;\;,
\end{equation}  
\end{small}
\vspace{+1pt}

\noindent
where $R^{(i)} : \allstate \times \allaction \rightarrow \mathbb{R}$ is an example-specific reward function, $H(\cdot)$ is an entropy regularization term, and $\lambda$ is the regularization coefficient. 
The reward function $R^{(i)}$ details are described in details in Appendix~\ref{apx:reward}. 
Roughly speaking, the reward function includes two additive components: a problem reward and a shaping term~\cite{Ng:99rewardshaping}. 
The problem reward provides a positive reward for successful task completion, and a negative reward for incorrect completion or collision. 
The shaping term is positive when the agent gets closer to the goal position, and negative if it is moving away. 
The gradient of the objective is:

\begin{small}
\begin{eqnarray*}
  \nabla J &=& \sum_{a\in \allaction} \pi(\action\mid \ostate_t) \nabla \log \pi(\action\mid \ostate_t) R(\state_t, \action) \\ 
	&& \hspace{2em} + \lambda \nabla H(\pi(. \mid \ostate_t)\;\;.
\end{eqnarray*} 
\end{small}

\noindent
We approximate the gradient by sampling an action using the policy~\cite{Williams:92reinforce}, and use the gold goal location computed from $\goalstate^{(i)}$. 
We perform several parallel rollouts to compute gradients and update the parameters using Hogwild!~\cite{recht2011hogwild} and Adam learning rates.

\section{Tasks and Data}
\label{sec:data}

\begin{table}[t]
	\begin{center}
    \footnotesize
		\begin{tabular}{|l|c|c|}	
			\hline
			Dataset Statistic & \navdrone & \house \\
			\hline
			Number paragraphs & $6{,}000$ & $1{,}596$ \\
			Mean instructions per paragraph & $4{.}7$ & $7{.}70$\\
			Mean actions per instruction & $24{.}6$ & $54{.}5$ \\
			Mean tokens per instruction & $12{.}1$ & $8{.}4$ \\
			Vocabulary size & $2{,}292$ & $1{,}018$ \\
			\hline
		\end{tabular}
	\end{center}
	\caption{Summary statistics of the two corpora.}
	\label{tab:navdrone-corpus}
\end{table}

\subsection{\navdrone}

The goal of \navdrone is to evaluate how well an agent can follow navigation instructions. The agent task is to follow a sequence of instructions that specify a path in an environment with multiple landmarks. Figure~\ref{fig:examples} (left) shows an example instruction. 

The environment is a fenced, square, grass field. 
Each instance of the environment contains between 6--13 randomly placed landmarks, sampled from 63 unique landmarks. 
The agent can take four types of discrete actions: $\act{Forward}$, $\act{TurnRight}$, $\act{TurnLeft}$, and $\stopaction$. 
The field is of size 50$\times$50, the distance of the $\act{forward}$ action is 1.5, and the turn angle is 15$^\circ$.
The environment simulator is implemented in Unity3D. 
At each time step, the agent performs an action, observes a first person view of the environment as an RGB image, and receives a scalar reward. 
The simulator provides a socket API to control the agent and the environment. 

\begin{figure}
  \centering
  \frame{\includegraphics[width=0.85\linewidth,clip,trim=38 558 199 30]{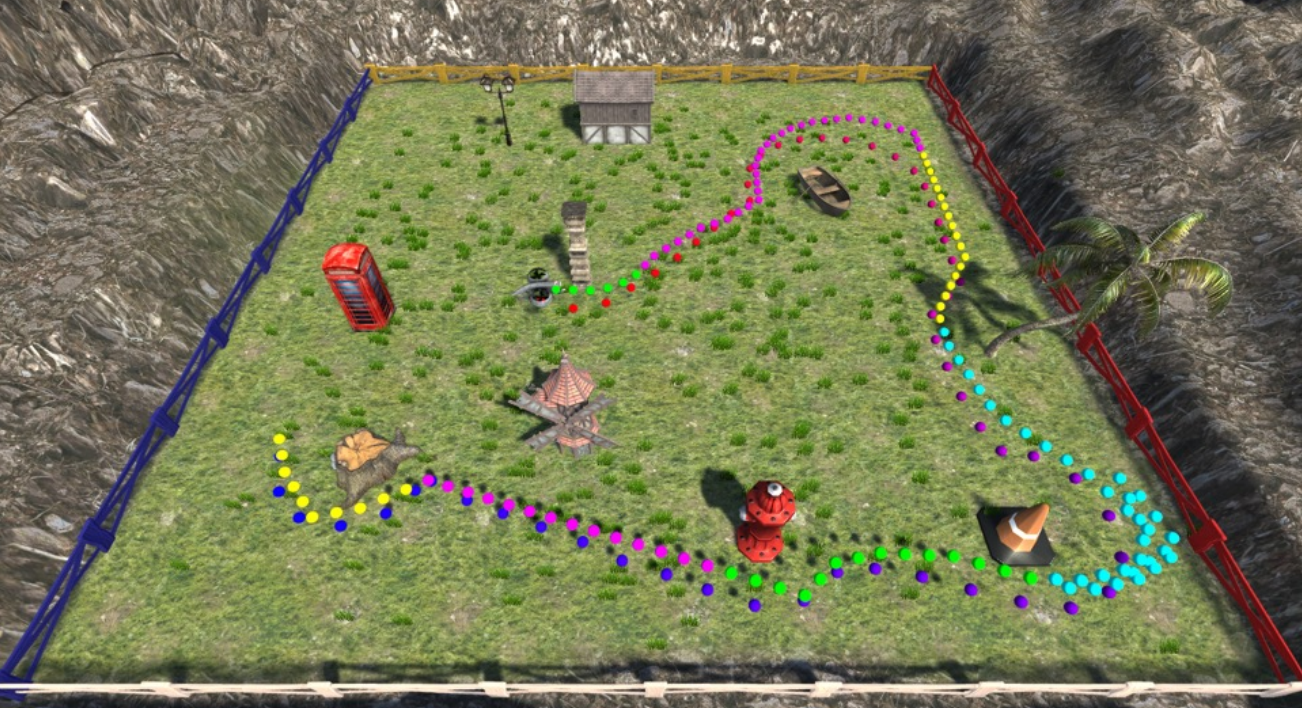}} \\[3pt]
  \footnotesize
  \begin{tabular}{|p{0.93\linewidth}|}
  \hline
  \textbf{[}\nlstring{\dronegreenuline{Go around the pillar on the right hand side}}\textbf{]} \textbf{[}\nlstring{\dronepinkuline{and head towards the boat, circling around it clockwise.}}\textbf{]} \textbf{[}\nlstring{\droneyellowuline{When you are facing the tree, walk towards it, and the pass on the right hand side,}}\textbf{]} \textbf{[}\nlstring{\dronecyanuline{and the left hand side of the cone. Circle around the cone,}}\textbf{]} \textbf{[}\nlstring{\dronegreenuline{and then walk past the hydrant on your right,}}\textbf{]} \textbf{[}\nlstring{\dronepinkuline{and the the tree stump.}}\textbf{]} \textbf{[}\nlstring{\droneyellowuline{Circle around the stump and then stop right behind it.}}\textbf{]} \\
  \hline
  \end{tabular}
  
  \caption{Segmented instructions in the \navdrone domain. The original reference path is marked in red (start) and blue (end). The agent, using a drone icon, is placed at the beginning of the path. The follower path is coded in colors to align to the segmented instruction paragraph.}
  \label{fig:collection:drone}
\end{figure}

Agent performance is evaluated using two metrics: task completion accuracy, and stop distance error. 
A task is completed correctly if the agent stops within an aerial distance of $5$ from the goal.
We collect a corpus of navigation instructions using crowdsourcing. 
We randomly generate  environments, and generate one reference path for each environment. 
To elicit linguistically interesting instructions, reference paths are generated to pass near landmarks. 
We use Amazon Mechanical Turk, and split the annotation process to two tasks. 
First, given an environment and a reference path, a worker writes an instruction paragraph for following the path.
The second task requires another worker to control the agent to perform the instructions and simultaneously mark at each point what part of the instruction was executed. 
The recording of the second worker creates the final data of segmented instructions and demonstrations. 
The generated reference path is displayed in both tasks. The second worker could also mark the paragraph as invalid. 
Both tasks are done from an overhead view of the environment, but workers are instructed to provide instructions for a robot that observes the environment from a first person view. 
Figure~\ref{fig:collection:drone} shows a reference path and the written instruction.
This data can be used for evaluating both executing sequences of instructions and single instructions in isolation. 

\begin{table*}[t]
	\begin{center}
		\footnotesize
		\begin{tabular}{|p{3cm}|c|c|L{9.2cm}|}
          \hline
		 & \multicolumn{2}{c|}{Count} &  \\
		 \cline{2-3}
         Category & \navdrone & \house & Example \\
         \hline
		 \multirow{2}{3cm}{Spatial relations between locations} & \multirow{2}{*}{123} & \multirow{2}{*}{52} & \navdrone: \nlstring{go to the \textbf{right side of the} rock} \\
          & & & \house: \nlstring{pick up the cup \textbf{next to the} bathtub and place it on \dots}\\
        \dline
		\multirow{2}{3cm}{Conjunctions of two more locations} & \multirow{2}{*}{36}	 & \multirow{2}{*}{5} & \navdrone: \nlstring{fly between \textbf{the mushroom and the yellow cone}} \\ 
        & & & \house: \nlstring{\dots set it on the table next to \textbf{the juice and milk}.}\\
        \dline
		\multirow{2}{3cm}{Temporal coordination of sub-goals} & \multirow{2}{*}{65}  & \multirow{2}{*}{68} & \navdrone: \nlstring{at the mushroom \textbf{turn right and move forward towards the statue}} \\
        & & & \house: \nlstring{\textbf{go back to} the kitchen \textbf{and   put the glass in the sink}.}\\
        \dline
		\multirow{2}{3cm}{Constraints on the shape of trajectory} & \multirow{2}{*}{94} & \multirow{2}{*}{0} & \multirow{2}{*}{\navdrone: \nlstring{go past the house \textbf{by the right side of the apple}}} \\
        & & & \\
        \dline
		\multirow{2}{3cm}{Co-reference} & \multirow{2}{*}{32} & \multirow{2}{*}{18} & \navdrone: \nlstring{turn around \textbf{it} and move in front of fern plant} \\
        & & & \house: \nlstring{turn left, towards the kitchen door and move through \textbf{it}.}\\
        \dline
		Comparatives & 2 & 0 & \navdrone: \nlstring{\dots the small stone \textbf{closest to the blue and white fences} stop} \\
		\hline
		\end{tabular}
	\end{center}
	\vspace{-2pt}
	\caption{Qualitative analysis of the \navdrone and \house corpora. We sample $200$ single development instructions from each corpora. For each category, we count how many examples of the $200$ contained it and show an example.}
	\label{tab:dataset-ling}
\end{table*}

Table~\ref{tab:navdrone-corpus} shows the corpus statistics.\footnote{Appendix~\ref{apx:corpora} provides  statistics for related datasets.} 
Each paragraph corresponds to a single unique instance of the environment. 
The paragraphs are split into train, test, and development, with a $70\%$ / $15\%$ / $15\%$ split. 
Finally, we sample $200$ single development instructions for qualitative analysis  of the language challenge the corpus presents (Table~\ref{tab:dataset-ling}).

\subsection{\house}

The \house corpus combines both navigation and simple manipulation in a complex, simulated household environment. 
We use the \chalet simulator~\cite{yan2018chalet}, a 3D house simulator that provides multiple houses, each with multiple rooms. The environment supports moving between rooms, picking and placing objects, and opening and closing cabinets and similar containers. Objects can be moved between rooms and in and out of containers. 
The agent observes the world in first-person view, and can take five actions: $\act{Forward}$, $\act{TurnLeft}$, $\act{TurnRight}$, $\stopaction$, and $\act{Interact}$. 
The $\act{Interact}$ action acts on objects. It takes as argument a 2D position in the agent's view. 
Agent performance is evaluated with two metrics: 
(a) stop distance, which measures the distance of the agent's final state to the final annotated position;
and (b) manipulation accuracy, which compares the set of manipulation actions to a reference set. 
When measuring distance, to consider the house plan, we compute the minimal aerial distance for each room that must be visited. 
\citet{yan2018chalet} provides the full details of the simulator and evaluation. 
We use five different houses, each with up to six rooms. Each room contains on average 30 objects. 
A typical room is of size 6$\times$6. 
We set the distance of $\act{forward}$ to 0.1, the turn angle to 90$^\circ$, and divide the agent's view to a 32$\times$32 grid for the $\act{Interact}$ action.

\begin{figure}
  \centering
  \footnotesize
  \begin{tabular}{|p{0.93\linewidth}|}
  \hline
  Scenario \\
  \hline
  \nlstring{You have several hours before guests begin to arrive for a dinner party. You are preparing a wide variety of meat dishes, and need to put them in the sink. In addition, you want to remove things in the kitchen, and bathroom which you don't want your guests seeing, like the soaps in the bathroom, and the dish cleaning items. You can put these in the cupboards. Finally, put the dirty dishes around the house in the dishwasher and close it.} \\
  \hline
  \hline
  Written Instructions \\
  \hline
  \textbf{[}\nlstring{In the kitchen, open the cupboard above the sink.}\textbf{]} \textbf{[}\nlstring{Put the cereal, the sponge, and the dishwashing soap into the cupboard above the sink.}\textbf{]}  \textbf{[}\nlstring{Close the cupboard.}\textbf{]}  \textbf{[}\nlstring{Pick up the meats and put them into the sink.}\textbf{]} \textbf{[}\nlstring{Open the dishwasher, grab the dirty dishes on the counter, and put the dishes into the dishwasher.}\textbf{]} \\
  \hline
  \end{tabular}
  \caption{Scenario and segmented instruction from the \house corpus.}
  \label{fig:collection:house}
\end{figure}

We collected a corpus of navigation and manipulation instructions using Amazon Mechanical Turk. We created 36 common household scenarios to provide a familiar context to the task.\footnote{We observed that asking workers to simply write instructions without providing a scenario leads to combinations of repetitive instructions unlikely to occur in reality.} 
We use two crowdsourcing tasks. First, we provide workers with a scenario and ask them to write instructions. The workers are encouraged to explore the environment and interact with it. 
We then segment the instructions to sentences automatically. 
In the second task, workers are presented with the segmented sentences in order and asked to execute them. After finishing a sentence, the workers request the next sentence. The workers do not see the original scenario. 
Figure~\ref{fig:collection:house} shows a scenario and the written segmented paragraph.
Similar to \navdrone, \house data can be used for studying complete paragraphs and single instructions. 

Table~\ref{tab:navdrone-corpus} shows the corpus statistics.\footnote{The number of actions per instruction is given in the more fine-grained action space used during collection. To make the required number of actions smaller, we use the more coarse action space specified.}
The paragraphs are split into train, test, and development, with a 70\% / 15\% / 15\% split. 
Table~\ref{tab:dataset-ling} shows qualitative analysis of a sample of 200  instructions.

\section{Experimental Setup}
\label{sec:experiments}

\paragraph{Method Adaptations for \house}

We apply two modifications to our model to support intermediate goal for the \house instructions. 
First, we train an additional RNN to predict the sequence of intermediate goals given the instruction only. 
There are two types of goals: $\act{navigation}$, for action sequences requiring movement only and ending with the $\stopaction$ action; and $\act{interaction}$, for sequence of movement actions that end with an $\act{interact}$ action. 
For example, for the instruction \nlstring{pick up the red book and go to the kitchen}, the sequence of goals will be $\langle \act{interaction}, \act{navigation}, \act{navigation}\rangle$. 
This indicates the agent must first move to the object to pick it up via interaction, move to the kitchen door, and finally move within the kitchen. 
The process of executing an instruction starts with predicting the sequence of goal types. 
We call our model (Section~\ref{sec:model}) separately for each goal type.
The execution concludes when the final goal is completed. 
For learning, we create a separate example for each intermediate goal and train the additional RNN separately. 
The second modification is replacing the backward camera projection for inferring the goal location with ray casting to identify $\act{interaction}$ goals, which are often objects that are not located on the ground.

\paragraph{Baselines} We compare our approach against the following baselines: 
(a) \system{Stop}: Agent stops immediately; 
(b) \system{RandomWalk}: Agent samples actions uniformly until it exhausts the horizon or stops; 
(c) \system{MostFrequent:} Agent takes the most frequent action in the data, $\act{Forward}$ for both datasets, until it exhausts the horizon;
(d) \system{Misra17}: the approach of \citet{Misra:17instructions}; and
(e) \system{Chaplot18}: the approach of \citet{chaplot2017gated}.
We also evaluate goal prediction  and compare to the method of \citet{janner2017representation} and a \system{Center} baseline, which always predict the center pixel. 
Appendix~\ref{apx:baselines} provides baseline details.

\paragraph{Evaluation Metrics} 
We evaluate using the metrics described in Section~\ref{sec:data}: stop distance (SD) and task completion (TC) for \navdrone, and stop distance (SD) and manipulation accuracy (MA) for \house. 
To evaluate the goal prediction, we report the real distance of the predicted goal from the annotated goal and the percentage of correct predictions. We consider a goal correct if it is within a distance of $5.0$ for \navdrone and $1.0$ for \house. 
We also report human evaluation for \navdrone by asking raters if the generated path follows the instruction on a Likert-type scale of 1--5.
Raters were shown the generated path, the reference path, and the instruction. 

\paragraph{Parameters}
We use a horizon of $40$ for both domains.
During training, we allow additional $5$ steps to encourage learning even after errors. 
When using intermediate goals in \house, the horizon is used for each intermediate goal separately. 
All other parameters and detailed in Appendix~\ref{apx:hyperparameters}.

\section{Results}
\label{sec:results}

\begin{table}[t]
\begin{center}
\footnotesize
\begin{tabular}{|l|c|c|c|c|}
\hline
& \multicolumn{2}{c|}{\navdrone} & \multicolumn{2}{c|}{\house} \\
\cline{2-5} 
Method & SD &  TC & SD & MA \\
\hline
\hline
\system{Stop} & 15.37 & 8.20 & 2.99 & 37.53\\
\system{RandomWalk} & 14.80 & 9.66 & 2.99 & 28.96 \\
\system{MostFrequent} & 19.31 & 2.94 & 3.80 & 37.53 \\
\hline
\hline 
\system{Misra17} & 10.54 & 22.9 & 2.99 & 32.25\\
\system{Chaplot18} & 9.05 & 31.0 & 2.99 & 37.53 \\
Our Approach (OA) & \textbf{8.65} & \textbf{35.72} & \textbf{2.75} & 37.53 \\
\hline
\hline
OA w/o RNN & 9.21 & 31.30 & 3.75 &  37.43 \\
OA w/o Language & 10.65 & 23.02 & 3.22 & 37.53 \\
OA w/joint  & 11.54 & 21.76 & 2.99 & 36.90 \\
\hline
\hline
OA w/oracle goals & 2.13 & 94.60 & 2.19 & 41.07 \\
\hline
\end{tabular}
\end{center}
\caption{Performance on the development data.} 
\label{tab:results-dev}
\end{table}

\begin{table}[t]
\begin{center}
\footnotesize
\begin{tabular}{|l|c|c|c|c|}
\hline
& \multicolumn{2}{c|}{\navdrone} & \multicolumn{2}{c|}{\house} \\
\cline{2-5} 
Method & SD &  TC & SD & MA \\
\hline
\hline
\system{Stop} & 15.18 & 8.29 & 3.59 & 39.77\\
\system{RandomWalk} & 14.63 & 9.76 & 3.59 & 33.29\\
\system{MostFrequent} & 19.14 & 3.15 & 4.36 & 39.77 \\
\hline 
\hline
\system{Misra17} & 10.23 & 23.2 & 3.59 & 36.84\\
\system{Chaplot18} & 8.78 & 31.9 & 3.59 & 39.76 \\
Our Approach & \textbf{8.43} & \textbf{36.9} & \textbf{3.34} & \textbf{39.97}\\
\hline 
\end{tabular}
\vspace{-2pt}
\end{center}
\caption{Performance on the held-out test dataset.}
\vspace{7pt}
\label{tab:results-test}
\end{table}

\begin{table}[t]
\begin{center}
\footnotesize
\begin{tabular}{|l|c|c|c|c|}
\hline
& \multicolumn{2}{c|}{\navdrone} & \multicolumn{2}{c|}{\house} \\
\cline{2-5} 
Method & Dist &  Acc & Dist & Acc \\
\hline
 \system{Center} & 12.0 & 19.51 & 3.41 & 19.0 \\
\citet{janner2017representation} & 9.61 & 30.26  & 2.81 & 28.3  \\
Our Approach & 8.67 & 35.83  & 2.12  & 40.3  \\
\hline 
\end{tabular}
\end{center}
\vspace{-2pt}
\caption{Development goal prediction performance. We measure distance (Dist) and accuracy (Acc).} 
\label{tab:results-goal-prediction}
\end{table}

\begin{table}[t]
\begin{center}
\footnotesize
\begin{tabular}{|l|c|c|c|}
\hline
Category & Present & Absent & $p$-value \\
\hline
Spatial relations & 8.75 & 10.09 & .262 \\
Location conjunction & 10.19 & 9.05 & .327 \\
Temporal coordination & 11.38 & 8.24 & .015 \\
Trajectory constraints & 9.56 & 8.99 & .607 \\
Co-reference & 12.88 & 8.59 & .016 \\
Comparatives & 10.22 & 9.25 & .906 \\
\hline
\end{tabular}
\end{center}
\vspace{-3pt}
\caption{Mean goal prediction error for \navdrone instructions with and without the analysis categories we used in Table~\ref{tab:dataset-ling}. The $p$-values are from two-sided $t$-tests comparing the means in each row.}
\vspace{+7pt}
\label{tab:linguistic-results}
\end{table}

Tables~\ref{tab:results-dev} and~\ref{tab:results-test} show development and test results. 
Both sets of experiments demonstrate similar trends. 
The low performance of \system{Stop}, \system{RandomWalk}, and \system{MostFrequent} demonstrates the challenges of both tasks, and shows the tasks are robust to simple biases. 
On \navdrone, our approach outperforms \system{Chaplot18}, improving task completion (TC) accuracy by $5\%$, and both methods outperform \system{Misra17}. 
On \house, \system{Chaplot18} and \system{Misra17} both fail to learn, while our approach shows an improvement on stop distance (SD). 
However, all models perform poorly on \house, especially on manipulation (MA). 

To isolate navigation performance on \house, we limit our train and test data to instructions that include navigation actions only. 
The \system{Stop} baseline on these instructions gives a stop distance (SD) of 3.91, higher than the average for the entire data as these instructions require more movement. Our approach gives a stop distance (SD) of 3.24, a 17\% reduction of error, significantly better than the 8\% reduction of error over the entire corpus. 

We also measure human performance on a sample of 100 development examples for both tasks. 
On \navdrone, we observe a stop distance error (SD) of 5.2 and successful task completion (TC) 63\% of the time. On \house, the human distance error (SD) is 1.34 and the manipulation accuracy is 100\%. 
The imperfect performance demonstrates the inherent ambiguity of the tasks. The gap to human performance is still large though, demonstrating that both tasks are largely open problems.

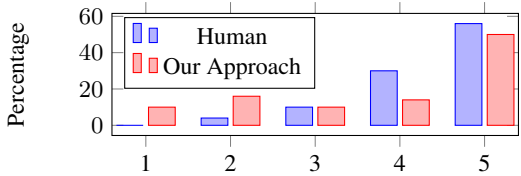
\begin{figure}[t]
\begin{tikzpicture}
    \footnotesize
    \begin{axis}[
            ybar,
            width=0.9\linewidth,
            height = 3.2cm,
            symbolic x coords={1,2,3,4,5},
            xtick=data,
            legend pos=north west,
            ylabel=Percentage,
        ]
        \addplot table[x=score,y=human, col sep=comma]{likert_data_binned.csv};
        \addplot table[x=score,y=model, col sep=comma]{likert_data_binned.csv};
        \legend{Human, Our Approach}
    \end{axis}
\end{tikzpicture}
\vspace{-2pt}
\caption{Likert rating histogram for expert human follower and our approach for \navdrone.}
\label{fig:humanlikert}
\end{figure}

The imperfect human performance raises questions about automated evaluation. In general, we observe that often measuring execution quality with rigid goals is insufficient.
We conduct a human evaluation with 50 development examples from \navdrone rating human performance and our approach. 
Figure~\ref{fig:humanlikert} shows a histogram of the ratings. 
The mean rating for human followers is 4.38, while our approach's is 3.78; we observe a similar trend to before with this metric.
Using judgements on our approach, we correlate the human metric with the SD measure. We observe a Pearson correlation -0.65 (p=5e-7), indicating that our automated metric  correlates  well with human judgment.\footnote{We did not observe this kind of clear anti-correlation comparing the two results for human performance (Pearson correlation of 0.09 and p=0.52). The limited variance in human performance makes correlation harder to test.}
This initial study suggests that our automated evaluation is appropriate for this task. 

Our ablations (Table~\ref{tab:results-dev}) demonstrate the importance of each of the components of the model. 
We ablate the action generation RNN (w/o RNN), completely remove the language input (w/o Language), and train the model jointly (w/joint Learning).\footnote{Appendix~\ref{apx:baselines} provides the details of joint learning.}
On \house especially, ablations results in models that display ineffective behavior. 
Of the ablations, we observe the largest benefit from decomposing the learning and using supervised learning for the language problem.

We also evaluate our approach with access to oracle goals (Table~\ref{tab:results-dev}). 
We observe this improves navigation performance significantly on both tasks. However, the model completely fails to learn a reasonable manipulation behavior for \house. This illustrates the planning complexity of this domain. 
A large part of the improvement in measured navigation behavior is likely due to eliminating much of the ambiguity the automated metric often fails to capture. 

\begin{figure}[t]
\begin{center}
\frame{\includegraphics[width=0.43\textwidth]{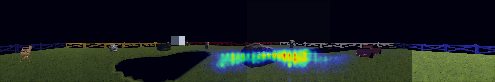}}
\fbox{\textit{\small curve around big rock keeping it to your left .}}\hfill\\
\vspace*{3pt}
\frame{\includegraphics[width=0.43\textwidth]{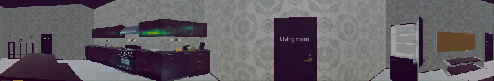}}
\fbox{\textit{\small walk over to the cabinets and open the cabinet doors up}}
\end{center}
\vspace{-5pt}
\caption{Goal prediction probability maps $\goalprob$ overlaid on the  corresponding observed panoramas $\pano$. The top example shows a result on $\navdrone$, the bottom on $\house$.}
\label{fig:goalpred_mini}
\end{figure}

Finally, on goal prediction (Table~\ref{tab:results-goal-prediction}), our approach outperforms the method of \citet{janner2017representation}. 
Figure~\ref{fig:goalpred_mini} and Appendix Figure~\ref{fig:goalpred} show example goal predictions.
In Table~\ref{tab:linguistic-results}, we break down \navdrone goal prediction results for the analysis categories we used in Table~\ref{tab:dataset-ling} using the same sample of the data. Appendix~\ref{apx:house-analysis} includes a similar table for \house.
We observe that our approach finds instructions with temporal coordination or co-reference challenging.  
Co-reference is an expected limitation; with single instructions, the model can not resolve references to previous instructions.

\vspace{-5pt}
\section{Discussion}
\label{sec:discussion}
\vspace{-3pt}

We propose a model for instruction following with explicit separation of goal prediction and action generation. 
Our representation of goal prediction is easily interpretable, while not requiring the design of logical ontologies and symbolic representations. 
A potential limitation of our approach is cascading errors. Action generation relies completely on the predicted goal and is not exposed to the language otherwise. 
This also suggests a second related limitation: the model is unlikely to successfully reason about instructions that include constraints on the execution itself. While the model may reach the final goal correctly, it is unlikely to account for the intermediate trajectory constraints. As we show (Table~\ref{tab:dataset-ling}), such instructions are common in our data. 
These two limitations may be addressed by allowing action generation access to the instruction. 
Achieving this while retaining an interpretable goal representation that clearly determines the execution is an important direction for future work. 
Another important open question concerns automated evaluation, which remains especially challenging when instructions do not only specify goals, but also constraints on how to achieve them. 
Our resources provide the platform and data to conduct this research.

\vspace{-5pt}
\section*{Acknowledgments}
\vspace{-3pt}

This research was supported by NSF (CAREER-1750499), Schmidt Sciences, a Google Faculty Award, and cloud credits from Microsoft. 
We thank John Langford, Claudia Yan, Bharath Hariharan, Noah Snavely, the Cornell NLP group, and the anonymous reviewers for their advice. 

\bibliographystyle{acl_natbib_nourl}
\bibliography{main}

\clearpage

\appendix
\newpage

\section{Tasks and Data: Comparisons}
\label{apx:corpora}

Table~\ref{tab:corpora} provides summary statistics comparing \navdrone and \house to existing related resources.

\begin{table*}
    \centering
    \footnotesize
    \begin{tabular}{|L{2cm}|c|c|c|c|c|c|}
        \hline
        Dataset & Num & Vocabulary & Mean Instruction   & Num. & Avg Trajectory   & Partially  \\
        		  & Instructions & Size & Length   & Actions & Length  & Observed   \\
        \hline
        \hline
        \citet{Bisk:16dataset} & 16,767& 1,426 & 15.27 & 81 & 15.4 & No  \\
        \dline
        \citet{MacMahon:06} & 3,237 & 563 & 7.96 &3 & 3.12 & Yes  \\
        \dline
        \citet{Matuszek:12b} & 217 & 39 & 6.65 & 3 & N/A & No  \\
        \dline
        \citet{Misra:15highlevel} & 469 & 775 & 48.7 & $>$100 & 21.5 & No   \\
        \hline
        \hline
        \navdrone & 28,204 & 2,292& 12.07 & 4 & 24.6 & Yes \\
        \dline
        \house & 13,729 & 1018 &  10.14 &  1028 & 54.5 & Yes \\
        \hline
    \end{tabular}
    \caption{Comparison of \navdrone and \house to several existing natural language instructions corpora.}
    \label{tab:corpora}
\end{table*}

\section{Reward Function}
\label{apx:reward}

\paragraph{\navdrone}

Following \citet{Misra:17instructions}, we use a shaped reward function that rewards the agent for moving towards the goal location. 
The reward for exampl $i$ is:

\begin{small}
\begin{equation}
  \reward^{(i)}(\state, \action, \state') = \reward_p^{(i)} + \phi^{(i)}(\state) - \phi^{(i)}(\state')
\end{equation}  
\end{small}

\noindent
where $\state'$ is the origin state, $\action$ is the action, $\state$ is the target state, $\reward_p^{(i)}$ is the problem reward, and $\phi^{(i)}(\state) - \phi^{(i)}$ is a shaping term. 
We use a potential-based shaping~\cite{Ng:99rewardshaping} that encourages the agent to both move and turn towards the goal. The potential function  is:

\begin{small}
\begin{eqnarray*}
\phi^{(i)}(\state) = & &\delta \textsc{TurnDist}(\state, \goalstate^{(i)}) \\ 
 \nonumber & & + (1 - \delta) \textsc{MoveDist}(\state, \goalstate^{(i)})\;\;,
\end{eqnarray*}
\end{small}

\noindent
where $\textsc{MoveDist}$ is the euclidean distance to the goal normalized by the agent's forward movement distance, and $\textsc{TurnDist}$ is the angle the agent needs to turn to face the goal normalized by the agent's turn angle. We use $\delta$ as a gating term, which is 0 when the agent is near the goal and increases monotonically towards 1 the further the agent is from the goal. This decreases the sensitivity of the potential function to the $\textsc{TurnDist}$ term  close to the goal.
The problem reward $\reward_p^{(i)}$ provides a negative reward of up to -1 on collision with any object or boundary (based on the angle and magnitude of collision), a negative reward of -0.005 on every action to discourage long trajectories, a negative reward of -1 on an unsuccessful stop, when the distance to the goal location is greater than 5, and a positive reward of +1 on a successful stop.

\paragraph{\house} We use a similar potential based reward function as \navdrone. Instead of rewarding the agent to move towards the final goal the model is rewarded for moving towards the next intermediate goal. We heuristically generate intermediate goals from the human demonstration by generating goals for objects to be interacted with, doors that the agent should enter, and the final position of the agent. The potential function  is: 

\begin{small}
\begin{eqnarray*}
\phi^{(i)}(\state) = & &\!\!\! \textsc{TurnDist}(\state, \state_{g,j}^{(i)}) + \\ 
&&\!\!\! \textsc{MoveDist}(\state, \state_{g,j}^{(i)}) + \textsc{IntDist}(\state, \state_{g,j}^{(i)})\;\;,
\nonumber
\end{eqnarray*}
\end{small}

\noindent
where $\state_{g,j}^{(i)}$ is the next intermediate goal, $\textsc{TurnDist}$ rewards the agent for turning towards the goal, $\textsc{MoveDist}$ rewards the agent for moving closer to the goal, and  $\textsc{IntDist}$ rewards the agent for accomplishing the interaction in the intermediate goal. The goal is updated on being accomplished. Besides the potential term, we use a problem  reward $\reward_p^{(i)}$ that gives a reward of 1 for stopping near a goal, -1 for colliding with obstacles, and -0.002 as a verbosity penalty for each step.

\section{Baseline Details}
\label{apx:baselines}

\paragraph{\system{Misra17}} We use the model of~\citet{Misra:17instructions}. The model uses a convolution neural network for encoding the visual observations, a recurrent neural network with LSTM units to encode the instruction, and a feed-forward network to generate actions using these encodings. The model is trained using policy gradient in a contextual bandit setting. We use the code provided by the authors.

\paragraph{\system{Chaplot18}} We use the gated attention architecture of~\citet{chaplot2017gated}. The model is trained using policy gradient with  generalized advantage estimation~\cite{schulman2015high}. We use the code provided by the authors. 

\paragraph{Our Approach with Joint Training} 
We train the full model with policy gradient. We maximize the expected reward objective with entropy regularization. Given a sampled goal location $l_g\sim p(.\mid \instruction, \pano)$ and a sampled action $a\sim p(.\mid \goalpos, (\image_1,\pose_1), \dots,(\image_{t}, \pose_{t}))$, the update is:

\begin{small}
\begin{eqnarray*}
  \nabla J &\approx& \left\{\nabla \log P(\goalpos \mid \instruction, \pano) \right. + \\ 
   && \left. \nabla \log P(\action_t \mid \goalpos, (\image_1,\pose_1), \dots,(\image_{t}, \pose_{t}) ) \right\}R(\state_t, \action) \\ 
	&& \hspace{0.2em}\lambda \nabla H(\pi(. \mid \ostate_t)\;\;.
\end{eqnarray*} 
\end{small}

\noindent
We perform joint training with randomly initialized goal prediction and action generation models. 

\section{Hyperparameters}
\label{apx:hyperparameters}

For \navdrone experiments, we use 5\% of the training data for tuning the hyperparameters and train on the remaining. For \house, we use the development set for tuning the hyperparameters. We train our models for 20 epochs and find the optimal stopping epoch using the tuning set. We use 32 dimensional embeddings for words and time. $\lstm_x$ and $\lstm_A$ are single layer LSTMs with 256 hidden units. The first layer of $\conv_0$ contains 128 8$\times$8 kernels with a stride of 4 and padding 3, and the second layer contains 64 3$\times$3 kernels with a stride of 1 and padding 1. The convolution layers in  \lingunet use 32 5$\times$5 kernels with stride 2. All deconvolutions except the final one, also use 32 5$\times$5 kernels with stride 2. The dropout probability in \lingunet is 0.5. The size of attention mask is 32$\times$32 + 1. For both $\navdrone$ and $\house$, we use a camera angle of 60$^\circ$ and create panoramas using 6 separate RGB images. Each image is of size 128$\times$128. We use a learning rate of 0.00025 and entropy coefficient $\lambda$ of 0.05.

\section{\house Error Analysis}
\label{apx:house-analysis}

Table~\ref{tab:house-linguistic-results} provides the same kind of error analysis results here for the \house dataset as we produced for \navdrone, comparing performance of the model on samples of sentences with and without the analysis phenomena that occurred in \house. 

\begin{table}
\begin{center}
\footnotesize
\begin{tabular}{|l|c|c|c|}
\hline
Category & Present & Absent & $p$-value \\
\hline
Spatial relations & 2.56 & 1.77 & .023 \\
Location conjunction & 3.85 & 1.93 & .226 \\
Temporal coordination & 1.70 & 2.14 & .164 \\
Co-reference & 1.98 & 1.98 & .993 \\
\hline
\end{tabular}
\end{center}
\caption{Mean goal prediction error for \house instructions with and without the analysis categories we used in Table~\ref{tab:dataset-ling}. The $p$-values are from two-sided $t$-tests comparing the means in each row.}
\label{tab:house-linguistic-results}
\end{table}

\section{Examples of Generated Goal Prediction}
\label{apx:goal-prediction-examples}

\begin{figure*}
\begin{center}
\frame{\includegraphics[width=\textwidth]{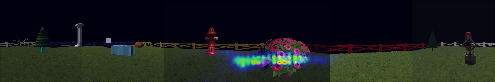}}
\startarrow{9.3}{0.48}
\textoverlay{14.94}{0.75}{Success}\hfill
\fbox{\textit{go round the flowers}}\hfill
\vspace*{4pt}

\frame{\includegraphics[width=\textwidth]{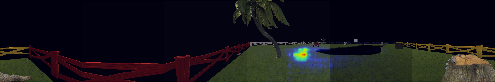}}
\startarrow{9.3}{0.48}
\textoverlay{14.94}{0.75}{Success}\hfill
\fbox{\textit{fly between the palm tree and pond}}\hfill
\vspace*{4pt}

\frame{\includegraphics[width=\textwidth]{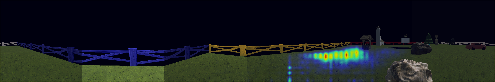}}
\startarrow{9.3}{0.48}
\textoverlay{15.00}{0.75}{Failure}\hfill
\fbox{\textit{head toward the wishing well and keep it on your right .}}\hfill
\vspace*{4pt}

\frame{\includegraphics[width=\textwidth]{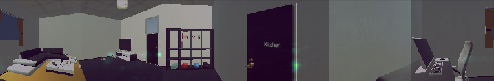}}
\startarrow{9.3}{0.48}\hfill
\fbox{\textit{move back to the kitchen .}}\hfill
\vspace*{4pt}

\frame{\includegraphics[width=\textwidth]{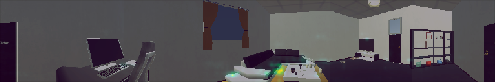}}
\startarrow{9.3}{0.48}\hfill
\fbox{\textit{then drop the tropicana onto the coffee table .}}\hfill
\vspace*{4pt}

\frame{\includegraphics[width=\textwidth]{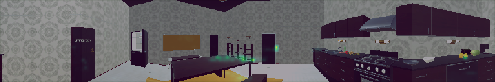}}
\startarrow{9.3}{0.48}\hfill
\fbox{\textit{walk the cup to the table and set the cup on the table .}}\hfill
\vspace*{1pt}

\end{center}
\caption{Goal prediction probability maps $\goalprob$ overlaid on the  corresponding observed panoramas $\pano$. The top three examples show results from $\navdrone$, the bottom three from $\house$. The white arrow indicates the forward direction that the agent is facing. The success/failure in the \navdrone examples indicate if the task was completed accurately or not following the task completion (TC) metric.}
\label{fig:goalpred}
\end{figure*}

Figure~\ref{fig:goalpred} shows example goal predictions from the development sets. 
We found the predicted probability distributions to be reasonable even in many cases where the agent failed to successfully complete the task. 
We observed that often the evaluation metric is too strict for \navdrone instructions, especially in cases of instruction ambiguity.

\end{document}